# A SUPERVISED INFORMATION ENHANCED MULTI-GRANULARITY CONTRASTIVE LEARNING FRAMEWORK FOR EEG BASED EMOTION RECOGNITION

*Xiang Li*[1,2], *Jian Song*[1,2], *Zhigang Zhao*[1,2,*], *Chunxiao Wang*[1,2,*], *Dawei Song*[3,*], *Bin Hu*[3]

[1] Key Laboratory of Computing Power Network and Information Security, Ministry of Education, Shandong Computer Science Center (National Supercomputer Center in Jinan), Qilu University of Technology (Shandong Academy of Sciences), Jinan, China
[2] Shandong Provincial Key Laboratory of Computer Networks, Shandong Fundamental Research Center for Computer Science, Jinan, China
[3] Institute of Engineering Medicine, Beijing Institute of Technology, Beijing, China

## ABSTRACT

This study introduces a novel **S**upervised **I**nfo-enhanced **C**ontrastive **L**earning framework for **E**EG based **E**motion **R**ecognition (**SI-CLEER**). SI-CLEER employs multi-granularity contrastive learning to create robust EEG contextual representations, potentially improving emotion recognition effectiveness. Unlike existing methods solely guided by classification loss, we propose a joint learning model combining self-supervised contrastive learning loss and supervised classification loss. This model optimizes both loss functions, capturing subtle EEG signal differences specific to emotion detection. Extensive experiments demonstrate SI-CLEER's robustness and superior accuracy on the SEED dataset compared to state-of-the-art methods. Furthermore, we analyze electrode performance, highlighting the significance of central frontal and temporal brain region EEGs in emotion detection. This study offers an universally applicable approach with potential benefits for diverse EEG classification tasks.

## 1. INTRODUCTION

Emotion recognition, a crucial aspect of Affective Computing, provides technology for intelligent emotional decoding and interactions. EEG-based emotion recognition, with its non-invasive, user-friendly, and high-time-resolution attributes, has made significant strides. However, challenges persist due to EEG's sensitivity to noise and low signal-to-noise ratio (SNR), thus decrease the reliability of emotion recognition systems [1, 2]. To overcome these challenges, there's a pressing need for developing robust feature extraction and spatio-temporal representation learning method that can effectively capture the distinctive neural signatures linked to various emotional states within EEG signals [3].

Zhigang Zhao (zhaozg@sdas.org), Chunxiao Wang (wangcx@sdas.org), Dawei Song (dwsong@bit.edu.cn) are corresponding authors. This work was supported by the Technology and Innovation Major Project of the Ministry of Science and Technology of China (grant No.2022ZD0118600), the Jinan "20 New Colleges and Universities" Funded Project (grant No.202333043), and the Talent Research Projects of Qilu University of Technology (Shandong Academy of Sciences) (grant No.2023RCKY136).

For this purpose, this study introduces a supervised information enhanced contrastive learning framework. By incorporating supervised classification information, it enhances the model's performance and generalization ability in emotion recogntion. The main contributions of this study are as follows: This study presents the first-ever end-to-end joint learning framework that merges self-supervised contrastive learning with supervised classification learning to acquire high-quality EEG representations for emotion recognition. Furthermore, this innovative approach is versatile and can be applied to a wide range of EEG-based intelligent analysis tasks beyond just emotion recognition. Based on this framework, we undertake an exploration of how various brain regions and electrodes influence classification accuracy. This effort provides valuable insights to enhance our understanding of the role of different EEG components in emotion recognition.

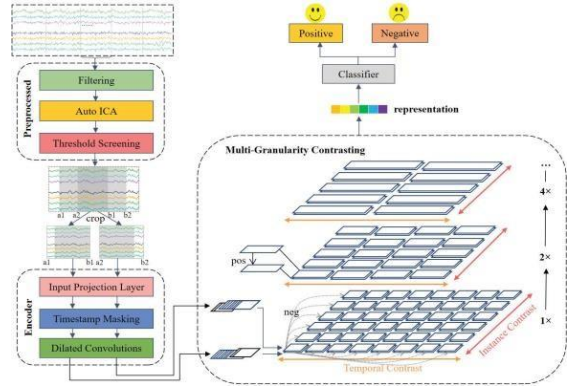

**Fig. 1**: The architecture diagram of the proposed model.

## 2. PROPOSED FRAMEWORK

### 2.1. Contrastive Representation Learning

Inspired by the idea of Contrastive Learning on time series [4], a self-supervised technique, our approach uses an encoder to create an unsupervised, non-parametric function $t = f(s)$, with $f$ being a deep neural network encoder specialized in extracting feature vectors from EEG signals. By training this encoder $f$, our model minimizes the distance between representations of an input sample $s$ and a similar one $s^+$ in the embedding space, while maximizing the distance from a dissimilar sample $s^-$ and these points. This concept is formalized in Equation (1), embedding a similarity metric. This metric empowers the model to distinguish various samples in the embedding space effectively, capturing commonalities and disparities among EEG signal samples. This approach results in more comprehensive, informative feature representations, enhancing performance in classification tasks.

$$similarity(f(s), f(s^+)) \gg similarity(f(s), f(s^-)). \quad (1)$$

## 2.2. Selection of Contextual Views

In the context of contrastive learning, the selection of positive and negative samples presents a significant challenge. To address this, we employ a dual strategy involving 'Timestamp Masking' and 'Random Cropping' on the input EEG signals. This approach generates a range of diverse contextual perspectives. Within these perspectives, we establish representations sharing the same timestamp as positive sample pairs, contrasting with representations carrying distinct timestamps that form negative sample pairs. This mechanism serves to actively guide the model in acquiring feature representations that effectively discriminate between analogous and dissimilar samples.

*1) Random Cropping:* Given a time series $x_i \in R^{T \times C}$, where $T$ denotes the number of timestamps and $C$ signifies the count of EEG channels, our approach involves the random cropping of two context views, $[a_1, b_1]$ and $[a_2, b_2]$. This cropping occurs along the time axis and adheres to the condition $0 < a_1 \leq a_2 \leq b_1 \leq b_2 \leq T$. Notably, the overlapping interval $[a_2, b_1]$ must exhibit consistent representations within both context views. This strategic choice ensures that the model grasps features that sustain coherence across these overlapping segments.

*2) Timestamp Masking:* For the output vector $z_i = \{z_{i,t}\}$ from the input projection layer, we introduce timestamp masking. We use a binary encoding vector $n \in \{0, 1\}^T$ to independently select mask positions, following a Bernoulli distribution with $p = 0.5$. Importantly, masking occurs in the latent vectors, not the original values. This approach ensures the process unfolds in a high-dimensional vector space, unconstrained by the original data value range. It also prevents the model from disproportionately focusing on specific timestamps, enhancing overall robustness.

## 2.3. Hierarchical and Multi-Granularity Contrast

This approach adopts the hierarchical and multi-granularity constrative learning to captures an array of feature insights across distinct time scales within EEG signals.

### 2.3.1. Dual Contrastive Learning on Single Semantic Level

The Dual Contrastive learning (DCL) Loss integrates Temporal Contrastive Learning (TCL) Loss and Instance-wise Contrastive Learning (ICL) Loss:

$$L_{DCL} = \frac{1}{NT} \sum_i \sum_t \left(l_{TCL}^{(i,t)} + l_{ICL}^{(i,t)}\right), \quad (2)$$

where $N$ is the total instance count, and $T$ signifies sample length in time steps. The details are as follows.

*1) Temporal Contrastive Learning:* In order to capture the evolving patterns within EEG signals over time, our model leverages the TCL. $z_{i,t}$ and $z'_{i,t}$ are positive pairs that denote the latent representations obtained from Encoder Module for the same timestamp $t$ but from two views of $x_{i,t}$. Conversely, representations originating from distinct timestamps are treated as negative pairs. The formulation of the temporal contrastive loss is presented below:

$$l_{TCL}^{(i,t)} = -\log \frac{\exp(z_{i,t} \cdot z'_{i,t})}{\sum_{t' \in \Omega}\left(\exp(z_{i,t} \cdot z'_{i,t'}) + \mathbb{1}_{[t \neq t']} \exp(z_{i,t} \cdot z_{i,t'})\right)}, \quad (3)$$

In this context, $\Omega$ signifies the collection of timestamps of the overlap part. The symbol $\mathbb{1}$ functions as an indicator function, taking the value 1 when $t \neq t'$ and 0 when $t = t'$.

*2) Instance-wise Contrastive Learning:* In our approach, we define an instance as a sample of length 2 seconds with a 0.2-second overlap. The latent representations obtained from Encoder Module for the same timestamp $t$ but from two views of $x_{i,t}$ are positive pairs. Whereas, the representation $z_{j,t}$ of diverse instances sharing the same timestamp $t$ within a given batch as negative samples. The formulation of the ICL is as follows:

$$l_{ICL}^{(i,t)} = -\log \frac{\exp(z_{i,t} \cdot z'_{i,t})}{\sum_{j=1}^{B}\left(\exp(z_{i,t} \cdot z'_{j,t}) + \mathbb{1}_{[i \neq j]} \exp(z_{i,t} \cdot z_{j,t})\right)}, \quad (4)$$

Here, $B$ denotes the batch size, and $j$ corresponds to the $j$-th EEG signal within the same batch. When $i \neq j$, $\mathbb{1}$ is set to 1; otherwise, it's set to 0. The notation $z_{j,t}$ signifies the representation of the $j$-th sample from the first view at timestamp $t$, while $z'_{j,t}$ denotes the representation of the $j$-th sample from the second view at the same timestamp $t$. These representations are integral in forming negative sample pairs for contrastive learning.

### 2.3.2. Contrastive Learning on Multiple Semantic Levels

The above DCL loss are calculated and recorded on various semantic (resolution) levels, namely multiple max-pooling operations are performed on the input series. After each max-pooling operation, the DCL loss is calculated. Since max-pooling operations decrease the length of the series, this process continues until the series length is reduced to 1. The overall Hierarchical Contrastive Learning (HCL) Loss $L_{\text{HCL}}$ is the average of the $L_{DCL}$ losses calculated after multiple max-pooling operations.

## 2.4. Model Architecture

Our model consists of four main components: the preprocessing module, the encoder module, the hierarchical contrastive module, and the downstream classifier, as shown in Figure 1. We will further describe these components in the subsequent sections.

*1) Preprocessing Module:* We utilize the MNE [5] toolkitfor raw data reading. EEG recordings are globally averaged according to the international 10-20 system standard. Bandpass filtering (1 to 49Hz) and notch filtering at 60Hz [6] effectively reduce noise. To address mixed activities and interferences from multiple sources in EEG signals, we apply Auto ICA [7] to eliminate artifacts. Importantly, this process doesn't necessitate manual intervention or expert knowledge.

*2) Encoder Module:* The encoder module is a crucial component in EEG signal processing. We selected the Input Projection Layer, Timestamp Masking Module, and Dilated CNN Module as the backbone of the encoder.

- **Input Projection Module:** The Input Projection Layer is a fully connected layer that maps the observed values $x_{i,t}$ at timestamp $t$ to a high-dimensional latent vector $z_{i,t}$. The purpose of this layer is to transform the original observed values into more meaningful and learnable representations.
- **Timestamp Masking Module:** Randomly masks vector representations, promoting robust feature learning. Enhanced context view generated by masking encourages location-insensitive feature representation learning. Avoiding focusing too much on specific timestamps, and improve the model's ability to generalize on unseen timestamps.
- **Dilated CNN Module:** The dilated CNN module comprises five residual blocks, with each block housing two one-dimensional convolutional layers. This module enhance the capture of long-range dependencies and context information within time series data.

*3) Hierarchical Contrastive Module:* Each parallelogram in Figure 1 depicts a representation vector at an instance's specific timestamp. This module captures multi-granularity contextual insights across time and instances via temporal and instance-wise contrastive learning as mentioned in Section 2.3.

*4) Classifier Module:* The downstream classifier comprises 1D convolutional, pooling, ReLU, and fully connected layers. While deeper networks often enhance learning performance [8], they can impact computation time. Hence, we meticulously crafted a network architecture for faster computation, and parameter fine-tuning during training achieves peak classification performance.

## 2.5. Model Optimization Objective

The literature [9] introduces a two-step supervised learning algorithm. However, training these components separately risks encoder representations ill-suited for downstream classification tasks. To address this, **we propose a joint training approach, optimizing self-supervised contrastive learning and supervised classification concurrently.** Guided by both contrastive and classification losses, contrastive learning adapts representations better suited for emotion detection. Specifically, we combine contrastive and classifier losses, as in equation (4):

$$Loss = L_{HCL} + L_{CLASS}. \quad (5)$$

For classification, we employ cross-entropy loss as $L_{class}$.

## 3. EXPERIMENT AND RESULTS

**Experimental Data and Setting:** The effectiveness of our framework is validated on the recognized SEED database [10]. The SEED database comprises EEG data obtained from 15 Chinese subjects using 62 EEG electrodes while they watched 15 Chinese film clips, each lasting approximately 4 minutes. Each subject participate the experiment 3 times on different days. Three distinct emotions—positive, neutral, and negative—were elicited. We segmented each participant's EEG data into segments of 2 seconds with a 0.2-second overlap.

The encoder's input dimension corresponds to the number of EEG channels, with a hidden dimension of 128, and an output representation dimension of 900. The encoder consists of 5 residual blocks in the hidden layers. We adopt the Adam optimizer with a learning rate of 0.001 and conduct training for 50 epochs in each cross-validation fold. ReLU activation functions were applied to all layers of the classifier, while the final layer utilized a Softmax activation function for three-emotion classification task.

**Table 1:** Detailed Recognition Performance on SEED

| Index  | Exp-1 ACC (%) | Exp-2 ACC (%) | Exp-3 ACC (%) |
|--------|---------------|---------------|---------------|
| Sub 01 | 93.13         | 96.82         | 99.43         |
| Sub 02 | 92.22         | 81.42         | 94.38         |
| Sub 03 | 95.11         | 93.69         | 89.49         |
| Sub 04 | 76.14         | 79.83         | 92.95         |
| Sub 05 | 96.93         | 98.86         | 97.84         |
| Sub 06 | 98.41         | 95.51         | 96.70         |
| Sub 07 | 99.20         | 95.74         | 98.81         |
| Sub 08 | 97.89         | 98.01         | 97.84         |
| Sub 09 | 95.85         | 98.97         | 95.00         |
| Sub 10 | 94.94         | 99.72         | 96.48         |
| Sub 11 | 94.89         | 95.34         | 94.94         |
| Sub 12 | 96.70         | 98.47         | 98.58         |
| Sub 13 | 98.75         | 99.89         | 99.20         |
| Sub 14 | 91.88         | 92.73         | 98.64         |
| Sub 15 | 99.66         | 99.20         | 99.20         |
|        |               | Mean          | 95.45         |

**Recognition Performance:** We employ a stratified K-fold cross-validation (SKCV) approach for assessing our model's performance on each subject's data. This method ensures a balanced sampling of three-type emotional samples. The results, averaged across 5 folds, are presented in Table 1. As demonstrated in Table 2, we present the performance of our SI-CLEER framework alongside that of several notable works with high Google citations. Our framework has improved emotion recognition performance on the SEED dataset.

**Table 2:** Performance Comparison

| Method | ACC (%) |
|---|---|
| 2D map of differential entropy feature combines with hierarchical convolutional neural network [11] [**Citations**: 231] | 88.20 |
| A dynamical graph convolutional neural network dynamically learns graph adjacency weight matrix and classifies emotions based on EEG features of DE, PSD, DASM, RASM and DCAU [12] [**Citations**: 728] | 90.40 |
| A bi-hemispheric discrepancy model(BiHDM) learns the asymmetric characteristics between hemispheres [13] [**Citations**: 131] | 93.10 |
| A method based on two deep generative models, variational autoencoder(VAE) and generative adversarial network (GAN), and two data augmentation strategies [14] [**Citations**: 82] | 93.50 |
| Attention-based LSTM with Domain Discriminator [15] [**Citations**: 91] | 91.10 |
| Contrastive learning without supervised information [16] [**Citations**: 90] | 85.70 |
| SI-CLEER (ours) | **95.45** |

The training loss curves for all experiments are depicted in Figure 2(left). Furthermore, as shown in Figure 2(reight) we employ the t-SNE method to visualize one subject's intermediate representations of EEGs from the SI-CLEER Encoder module in two-dimensional space. In the resulting visualization, green points denote neutral-emotion samples, while red points and blue points denote positive and negative emotion samples, respectively. Our framework effectively separates these samples into groups, with small intra-class distances among samples of the same class and large inter-class distances among samples of different classes. We can see that this intermediate representation is clearly advantageous for downstream classification tasks.

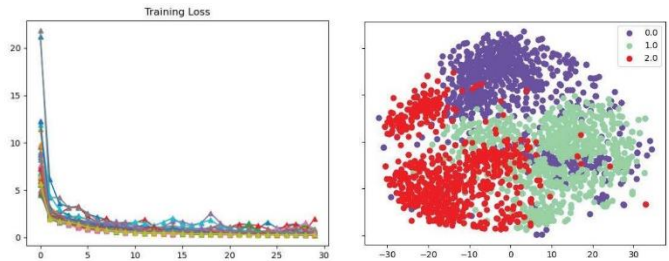

**Fig. 2:** Training loss curve of each experiments' data (left) and the t-SNE visulization of the intermediate representation from the encoder module of SI-CLEER(right).

**Exploration of Channel Importance:** The question of which channels and components in multi-channel EEG are advantageous for emotion recognition has long been a focal point of researchers' attention. Utilizing our framework, we evaluated emotion recognition performance across various channel data and visually presented the results through EEG topographic maps in Figure 3. It is evident that EEG electrodes that significantly contribute to emotion recognition are distributed primarily in the frontal central regions and bilateral temporal regions. This conclusion is consistent with findings from several studies in emotional-cognitive neuroscience [17, 18]. The top 5 recognition accuracy rankings ranges from 86.41% to 90.76%. Of course, we can see that each electrode contains useful information for emotion recognition. By utilizing information from all electrodes, an overall effectiveness of 95.45% can be achieved.

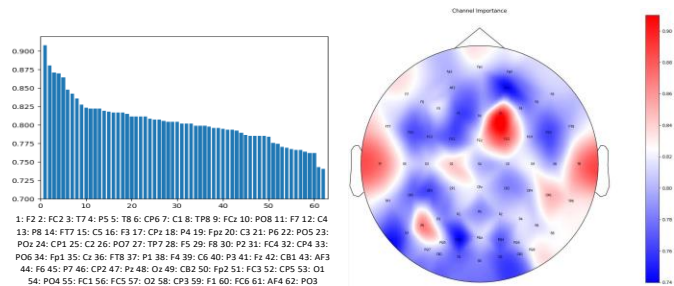

**Fig. 3:** Averaged performance (Accuracy) of each individual EEG channel in emotion recognition by SI-CLEER.

## 4. CONCLUSION

This paper introduces a novel approach that combines self-supervised contrastive learning with supervised classification for EEG-based emotion recognition. This joint training method improves supervision, promoting the acquisition of more informative features and addressing key challenges in EEG-based emotion recognition. Our method incorporates timestamp masking and random cropping to create diverse context views from input EEG signals. By utilizing contrastive learning with an encoder, we extract EEG signal representations, capturing multi-scale context through temporal and instance-wise contrastive learning. The proposed framework achieves enhanced accuracy on the benchmark SEED dataset compared with state-of-the-art EEG-based emotion recognition models. Notably, our results highlight the significance of central frontal and temporal brain regions in discrimination. Future directions may involve conducting experiments on more benchmark datasets and weighting the contribution of contrastive loss and classification loss in optimization. All source codes and data related to this study are freely accessible at https://github.com/muzixiang/SI-CLEER.